\pdfoutput=1

\documentclass[11pt]{article}
\usepackage[preprint]{acl}
\usepackage[normalem]{ulem}
\useunder{\uline}{\ul}{}
\usepackage{xcolor}
\usepackage{colortbl}
\definecolor{myblue}{RGB}{235,245,250}
\usepackage{times}
\usepackage{latexsym}
\usepackage{multirow}
\usepackage{graphicx}
\usepackage{amsmath}
\usepackage{amssymb}
\usepackage{booktabs}
\usepackage{subcaption}
\usepackage{caption}
\usepackage{bbding}
\usepackage{tcolorbox}

\newcommand{\ie}{\textit{i.e.}}
\newcommand{\eg}{\textit{e.g.}}
\usepackage{algorithm}
\usepackage{algorithmic}

\usepackage[T1]{fontenc}

\usepackage[utf8]{inputenc}

\usepackage{microtype}
\usepackage{bbold}

\usepackage{inconsolata}

\title{
\raisebox{-0.2cm}{\includegraphics[height=1cm]{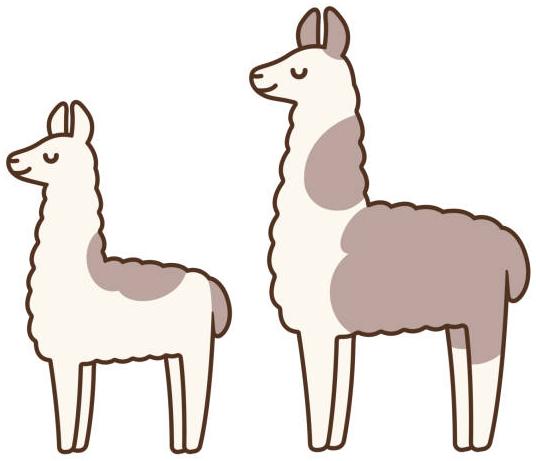}} CoRanking: Collaborative Ranking with Small and Large Ranking Agents
}

\author{Wenhan Liu$^1$, Xinyu Ma$^2$, Yutao Zhu$^1$, Lixin Su$^2$, \textbf{Shuaiqiang Wang}$^2$ \\ 
\textbf{Dawei Yin}$^2$ \and \textbf{Zhicheng Dou}$^{1}$\thanks{Corresponding author.} \\
$^1$Gaoling School of Artificial Intelligence, Renmin University of China \\
$^2$Baidu Inc., Beijing, China \\
\texttt{lwh@ruc.edu.cn, xinyuma2016@gmail.com, dou@ruc.edu.cn}
}

\begin{document}
\maketitle

\begin{abstract}
Listwise ranking based on Large Language Models (LLMs) has achieved state-of-the-art performance in Information Retrieval (IR).
However, their effectiveness often depends on LLMs with massive parameter scales and computationally expensive sliding window processing, leading to substantial efficiency bottlenecks. 
In this paper, we propose a Collaborative Ranking framework (\textbf{CoRanking}) for LLM-based listwise ranking.
Specifically, we strategically combine an efficient \emph{small} reranker and an effective \emph{large} reranker for collaborative ranking.
The \emph{small} reranker performs initial passage ranking, effectively filtering the passage set to a condensed top-k list (e.g., top-20 passages), and the \emph{large} reranker (with stronger ranking capability) then reranks only this condensed subset rather than the full list, significantly improving efficiency. 
We further address that directly passing the top-ranked passages from the small reranker to the large reranker is suboptimal because of the LLM's strong positional bias in processing input sequences. 
To resolve this issue, we propose a passage order adjuster learned by RL that dynamically reorders the top passages returned by the small reranker to better align with the large LLM's input preferences. 
Our extensive experiments across three IR benchmarks demonstrate that CoRanking achieves superior efficiency, reducing ranking latency by approximately 70\% while simultaneously improving effectiveness, compared to the standalone large reranker.

\end{abstract}

\begin{figure*}[t]
	\centering
        \vspace{-3mm}
	\includegraphics[width=0.85\linewidth]{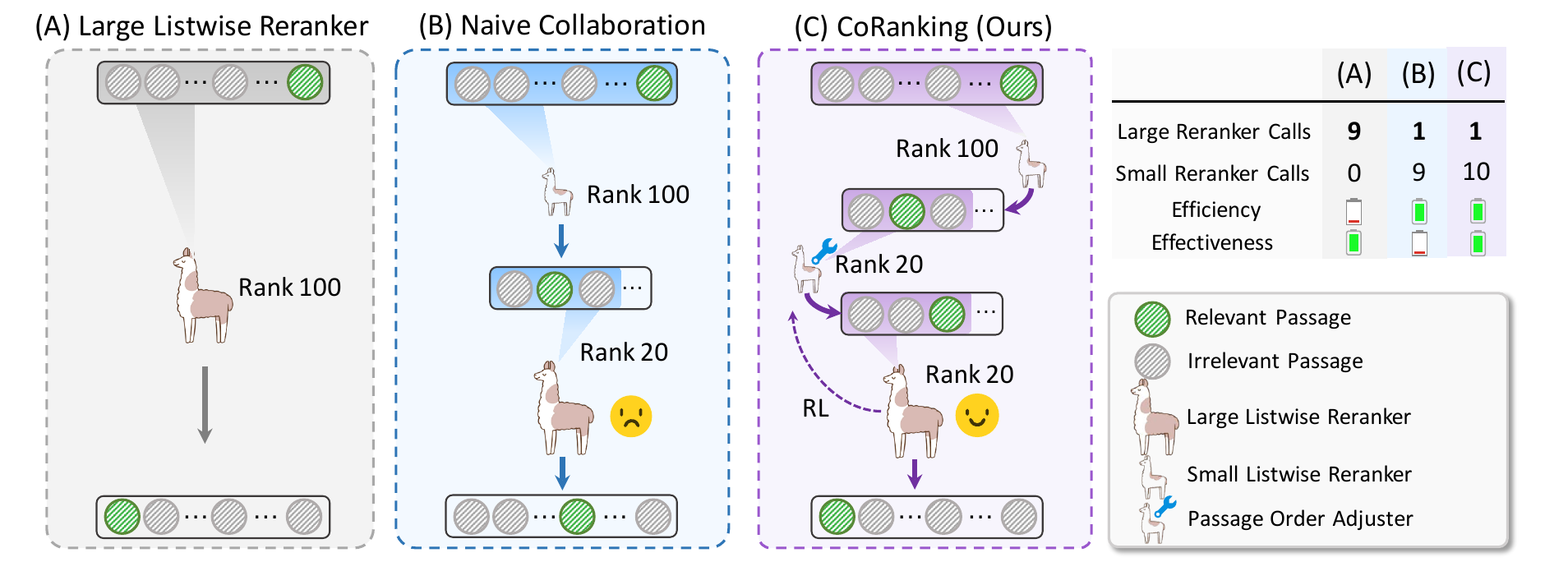}
	\caption{(A) Large Listwise Reranker: Apply a large listwise reranker to rerank 20 passages from 100 candidates based on a sliding window strategy. (B) Naive Collaboration: Pre-rank 100 passages with a small listwise reranker using a sliding window strategy, then rerank the top 20 with a large listwise reranker. (C) CoRanking: Pre-rank 100 passages with a small reranker using a sliding window strategy, adjust the order of the top 20 via the passage order adjuster, then rerank them with the large listwise reranker. ``Large/Small Reranker Calls'' represents the number of sliding windows for listwise ranking.}
	\label{fig:intro_example}
\end{figure*}

\section{Introduction}
\label{sec:intro}

In recent years, many studies~\cite{fullrank, rankgpt, llm4ir_survey} have demonstrated strong zero-shot passage ranking capabilities of large language models (LLMs). A typical approach is listwise ranking, which feeds the query and a list of passages into an LLM and instructs it to output a ranked list of passage IDs in descending order of relevance. This approach has established state-of-the-art performance~\cite{rankgpt} on major IR benchmarks, including TREC~\cite{dl19} and BEIR~\cite{beir}.

Despite its effectiveness, LLM listwise passage ranking faces critical efficiency challenges, as illustrated in Figure~\ref{fig:intro_example}(A), primarily stemming from two factors: First, its state-of-the-art performance relies on LLMs with massive parameters (such as Qwen2.5-72B or GPT-4), which introduces significant computational overhead and results in an efficiency issue. Second, due to the limitation of the context length of LLMs, existing listwise ranking methods typically employ a sliding window strategy~\cite{rankgpt}. With a fixed window size (e.g., 20 passages) and step size (e.g., 10 passages), this strategy bubbles the relevant passages that are initially lowly ranked to the top. However, the interdependence between windows hinders the parallelization of sliding windows, which results in many repetitive and sequential rankings of sliding windows (\eg, totally 9 sliding windows if window size and step size are set as 20 and 10 when ranking the top 20 from 100 passages) and creates an efficiency bottleneck (especially when employing LLMs with massive parameters).

The sliding window strategy of LLM reranker primarily serves to elevate initially low-ranked relevant passages into the final reranking window (\eg, top 20). We believe that this repetitive and inefficient process could be replaced by a small listwise reranker.\footnote{In this work, we consider models with parameters less than or equal to 3B as small rerankers.} By utilizing a small listwise reranker to pre-rank the whole list of passages (e.g., 100 passages) using a sliding window strategy, the large listwise reranker, which has stronger ranking capability, only needs to rerank the passages in the final window (\ie, containing the top 20 passages). This hybrid approach could dramatically reduce computational costs and improve efficiency.

However, existing small listwise rerankers are mainly trained on golden lists whose passages are ranked solely based on relevance~\cite{rankzephyr, rankvicuna}. We argue that directly feeding the top-ranked passages returned by these small listwise rerankers into a large listwise reranker may lead to sub-optimal performance (see Figure~\ref{fig:intro_example} (B)). This is because the effectiveness of a listwise reranker has a complex relationship with the order of input passages, rather than being determined exclusively by their relevance, as indicated by existing studies~\cite{rankgpt, self_consistency_llm}. For example, ~\citet{self_consistency_llm} reveal that relevant passages appearing in the middle of the list tend to be ranked at the bottom by the listwise reranker. This stems from LLMs' pre-training biases toward the positions of input tokens~\cite{LiuLHPBPL24}. While the passage order generated by the small listwise reranker, which is trained using relevance-based ranking, may deviate from the large listwise reranker's order preference. To address this issue, we introduce a passage order adjuster (also a small listwise reranker) to rerank the top-ranked passages from the small listwise reranker so that they align with the order preference of the large listwise reranker (see Figure~\ref{fig:intro_example} (C)).

More specifically, we propose a collaborative ranking framework, namely \textbf{CoRanking}, that integrates small and large ranking models to perform passage ranking in a multi-stage manner. In the first stage, we train a small listwise reranker (SLR) to pre-rank all the passages to elevate the relevant passages to the top part of the list (\eg, top-20). Different from existing methods~\cite{rankvicuna,rankzephyr} that solely use teacher-generated ranking lists as training labels, we propose a human-label-enhanced method to improve the ranking quality. In the second stage, we train a passage order adjuster (POA) using the reinforcement learning algorithm DPO. The POA reorders the top-ranked passages to align with the large listwise reranker's order preference. In addition, we design a significance-aware selection strategy $\text{S}^3$ to build high-quality preference pairs for DPO training. In the third stage, a large listwise reranker is used to further rerank the top-ranked passages, enhancing the overall ranking effectiveness. Extensive experiments on diverse IR benchmarks demonstrate that our CoRanking framework achieves a 70\% reduction in ranking latency while maintaining superior effectiveness, compared to a standalone large listwise reranker. 

Our contributions are summarized as follows:

(1) We propose a ranking framework, CoRanking, that enables small and large LLMs to collaborate, thereby achieving efficient and effective passage ranking.

(2) We propose a novel passage order adjuster (POA) that aligns the output passages of the small listwise reranker with the order preference of the large listwise reranker. We further design the $\text{S}^3$ strategy, which builds valuable preference pairs for POA's training.

(3) We conduct extensive experiments on diverse IR benchmarks, and the results demonstrate that our CoRanking achieves a significant efficiency improvement compared to a standalone large listwise reranker while exhibiting superior performance.

\begin{figure*}[!tb]
  \centering
  \includegraphics[width=1\linewidth]{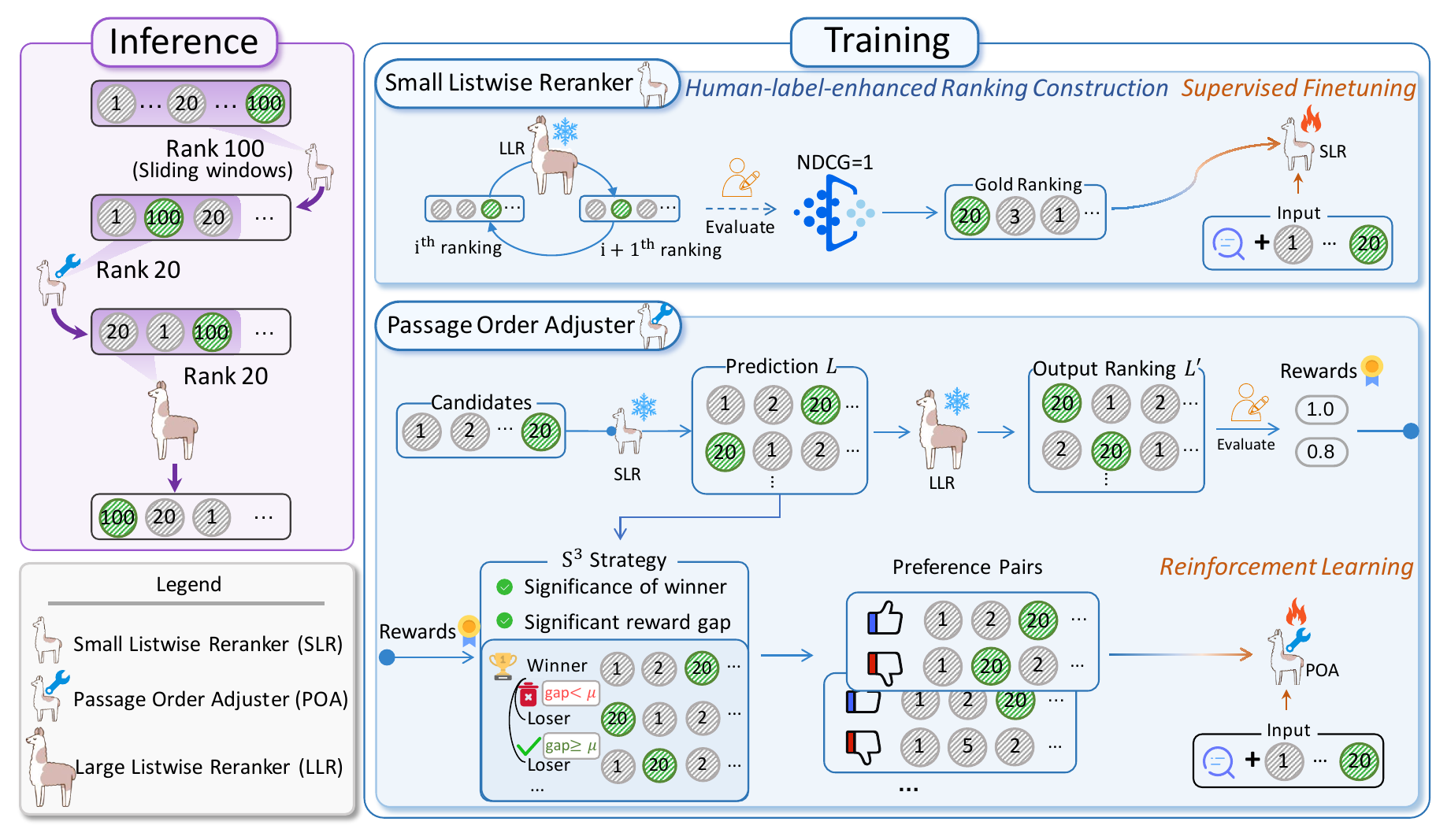}
  \caption{The ``Inference'' part shows how our CoRanking works with both large and small listwise rerankers. As for training of the small listwise reranker, we design a human-label-enhanced ranking construction method to generate gold rankings for supervised fine-tuning. As for training the passage order adjuster, we proposed the $\text{S}^3$ strategy to obtain high-quality preference pairs for DPO reinforcement learning.}
  \label{fig:model}
\end{figure*}

\section{Methodology}
\label{sec:method}

In this section, we present a collaborative ranking framework CoRanking that integrates both small and large passage rerankers. As illustrated in Figure~\ref{fig:model}, the framework comprises three key components: (1) a small listwise reranker, (2) a passage order adjuster, and (3) a large listwise reranker. Next, we will introduce the definition of the ranking task, the inference process of CoRanking, and the training details.

\subsection{Task Definition}
The aim of passage ranking is to rerank a list of retrieved passages $[p_1, \ldots, p_N]$ based on their relevance to a query $q$. The listwise ranking approach takes both $q$ and a set of passages as input, and generates a reranked sequence of passage IDs (\eg, [4] [2]\ldots). Due to the constraints of LLMs' input length, existing listwise ranking approaches typically employ a sliding window strategy that iteratively processes passage subsets. Specifically, this strategy applies a window of size $w$ that moves from the end to the beginning of the passage list with a step size $s$, dynamically promoting relevant passages to higher positions. Following previous studies~\cite{rankgpt}, we set the total number of passages $N$ to 100, with window size $w$ and step size $s$ set to 20 and 10, respectively.

\subsection{Inference Process of CoRanking}
Given the passages to rerank, our CoRanking (see Figure~\ref{fig:model}) first applies the small listwise reranker to pre-rank all the passages using a sliding window strategy. Then, we use the passage order adjuster to rerank the top-20 passages in a listwise manner, so that their order meets the preference of the large listwise reranker. Finally, the large listwise reranker is applied to further rerank the top-20 passages to obtain the final ranking list. This avoids the large listwise reranker's inefficient sliding window operations on all 100 passages, significantly improving ranking efficiency. The specific prompt template for a large listwise reranker is provided in Appendix~\ref{app:listwise_prompt}, which is consistent with previous studies~\cite{rankzephyr, fullrank}. In the following, we will illustrate the training details of our small listwise reranker and passage order adjuster.

\subsection{Small Listwise Reranker (SLR)}
To mitigate the latency and computational overhead of the large listwise reranker (denoted as LLR), we propose using a small listwise reranker (denoted as SLR) to pre-rank the entire passage list. The SLR aims to elevate relevant passages into the final sliding window (i.e., the top-20 passages), thereby reducing the number of LLR's sliding window iterations. Here we adopt listwise training because we find that it performs better than other approaches (\eg, pointwise training~\cite{monobert}), and the SFT-trained SLR naturally serves as the initialization for our listwise passage order adjuster, which will be detailed in Section~\ref{subsec:poa}.

\noindent\textbf{Training Label}\quad Conventional listwise reranker training typically employs knowledge distillation~\cite{rankvicuna,rankzephyr}, which uses a strong teacher model (\eg, GPT-4) to rerank BM25-retrieved top-20 passages, with the teacher-generated rankings serving as supervised fine-tuning (SFT) labels. However, the teacher model can not perform consistently well across all queries, such as long-tail queries, which can result in suboptimal performance. To further improve the SFT label, we propose a \textit{human-label-enhanced ranking construction} (HRC) method that improves teacher-generated rankings with human labels. 

Specifically, we first iteratively rerank BM25-retrieved top-20 passages $M$ times using a powerful teacher model in a listwise manner, where the $(i+1)$-th reranking input is the output ranking from the $i$-th iteration. In this paper, we apply LLR as our teacher model and set the maximum iteration number to 5. This progressive reranking approach is motivated by prior studies~\cite{rankzephyr,rankvicuna} which reveal that iteratively reranking enables gradual improvement of ranking lists. Then, we choose the ranking where human-annotated relevant passages are ranked at the top (\ie, NDCG=1) during the iteration process as the gold ranking.

\noindent\textbf{Training Loss}\quad After obtaining the gold ranking, we concatenate the query $q$ and the candidate passages $[p_1, \ldots, p_N]$ together as the training input $x$ and optimize our SLR by minimizing the standard language modeling loss $\mathcal{L}$:
\begin{align}
\label{equ:sft_loss}
\mathcal{L} &= -\sum_{i=1}^{|y|} \log(P_{\theta}(y_i \mid x, y_{<i})), \\
\label{equ:input_format}
x&=q \circ \text{[1]} \circ p_1 \circ \text{[2]} \circ p_2 \ldots \text{[N]} \circ p_N,
\end{align}
where $[i]$ represents the ID of passage $p_i$ and $y$ is the gold ranking label (\eg, ``[4] [2] \ldots'').

\subsection{Passage Order Adjuster (POA)} \label{subsec:poa}
While our SLR is trained to rank relevant passages at the top of the list, the order of top-ranked passages (\ie, the top-20 passages) may not align well with the order preference of LLR. To address this issue, we employ a reinforcement learning (RL) algorithm to train a passage order adjuster POA based on LLR's feedback to rerank these top-ranked passages. We adopt the Direct Preference Optimization (DPO)~\cite{dpo} algorithm for its optimization stability. The following sections elaborate on the construction of DPO training data and model optimization.

\noindent\textbf{Training Data}\quad The training data of DPO consists of a series of prediction pairs. Each prediction pair contains a winner and a loser prediction, namely, a passage ID list. The reward $R$ for each prediction $L$ is computed by feeding $L$ to LLR and evaluating the ranking metric $M$ of \textit{LLR's output ranking $L'$} based on human-annotated relevance label $y$:
\begin{equation}
\label{equ:reward}
R=M(L', y),
\end{equation}
where we use NDCG@10 as our metric $M$. 

We apply our trained SLR to rerank BM25-retrieved top-100 passages with a sliding window strategy for training queries and use the top-20 reranked passages as training candidate passages, which aims to align with the input passage distribution encountered by POA during inference. To generate the preference pairs for training POA, we first sample $m$ predictions from SLR by taking the candidates as the input, then evaluate these predictions through Equation~\ref{equ:reward}. After that, the highest-reward list is selected as the winner prediction, while others serve as loser predictions. Next, we pair the winner prediction with all the loser predictions to construct initial training preference pairs. 

To ensure the quality of these preference pairs, we introduce a \textit{significance-aware selection strategy} $\text{S}^3$ to filter out low-quality pairs. Specifically, the $\text{S}^3$ strategy follows two rules: (1) \textit{Significance of Winner}: The reward of the winner prediction must be 1 and exceed LLR’s ranking metric (\ie, the NDCG@10 of directly use LLR to rerank the candidate passages), ensuring the winner prediction provides positive improvement for final reranking of LLR; (2) \textit{Significant Gap between winner and loser}: The reward difference between winner prediction and loser prediction must exceed a threshold $\mu$, eliminating noisy pairs with marginal differences. 

\noindent\textbf{Optimization}\quad Given constructed preference pairs, we optimize POA using the following DPO objective:
\begin{align}
\mathcal{L}_{\text{DPO}} &= -\underset{x, y_w, y_l}{E} \left[ \log \sigma \left( \beta \log \frac{\pi_\theta^w \pi_f^t}{\pi_f^w \pi_\theta^t} \right) \right], \\
\pi_\theta^w &= \pi_\theta(y_w \mid x), \quad \pi_f^t = \pi_f(y_l \mid x),\\
\pi_f^w &= \pi_f(y_w \mid x), \quad \pi_\theta^t = \pi_\theta(y_l \mid x), 
\end{align}
where $x$ represents the training input whose format is the same as Equation~(\ref{equ:input_format}). $y_w$, $y_l$ represent the winner prediction and loser prediction, respectively. $\pi_\theta$ is the policy model to be optimized, and $\pi_f$ is the original policy model, which is initialized with our trained SLR. The frozen $\pi_f$ acts as a regularizer to maintain generation stability during optimization, and $\beta$ balances preference alignment and regularization towards $\pi_f$.

\section{Experiments}
\subsection{Setting}
\paragraph{Evaluation Datasets} \label{data}
We evaluate our framework on three established information retrieval benchmarks: TREC DL~\cite{dl19}, BEIR~\cite{beir}, and BRIGHT~\cite{bright}. For the TREC DL benchmark, we adopt the TREC Deep Learning 2019 (DL19)~\cite{dl19} and 2020 (DL20)~\cite{dl20} datasets derived from MS MARCO v1. The BEIR benchmark~\cite{beir} is selected to evaluate cross-domain generalization capabilities. We choose three representative datasets: TREC-Covid, Robust04, and Trec-News. BRIGHT~\cite{bright} is a reasoning-intensive IR benchmark whose relevance assessment transcends simplistic keyword matching or semantic similarity, instead demanding intentional and deliberate reasoning. We choose Economics, Earth Science, and Robotics datasets for evaluation. The detailed descriptions of each dataset will be provided in Appendix~\ref{app:datasets}.

\paragraph{Implementation Details}
As for the supervised fine-tuning of our SLR, we use Qwen2.5-3B-Instruct as the backbone model. The training queries consist of a total of 2k queries sampled from the MS MARCO training set. We utilize Qwen2.5-72B-Instruct\footnote{\url{https://huggingface.co/Qwen/Qwen2.5-72B-Instruct-GPTQ-Int4}} as our teacher model to SFT our SLR and also employ it as the LLR model throughout this study unless otherwise specified. As for the RL training of POA, we use our trained SLR for model initialization. We set the threshold $\mu$ as 0.4 and generate about 12k preference pairs from 2k queries for the DPO algorithm. Please see the Appendix~\ref{app:implementation} for more implementation details.

\paragraph{Baselines}
We compare our framework against a series of baselines, which can be divided into two parts: (1) Single reranker, (2) Collaborative rerankers (\ie reranking with small and large rerankers).

As for single reranker, we compare our trained SLR with a series of state-of-the-art supervised rerankers, including \textbf{monoBERT (340M)}~\cite{monobert}, \textbf{monoT5 (3B)}~\cite{monot5}, \textbf{RankT5 (3B)}~\cite{rankt5}, \textbf{RankVicuna (7B)}~\cite{rankvicuna}, \textbf{RankZephyr (7B)}~\cite{rankzephyr} and $\textbf{RankMistral}_{\text{100}}$~\cite{fullrank}. The first three are pointwise rerankers trained from human labels in the MS MARCO dataset~\cite{MSMARCO}. The last three are listwise rerankers which are distilled from GPT-3.5, GPT-4, and GPT-4o, respectively. To ensure more fair comparison with SLR in model architecture and training data, we introduce baseline \textbf{Qwen-pointwise}, which is a pointwise reranker based on Qwen2.5-3B-Instruct. This model concatenates a query-passage pair as input and outputs a relevance score. We utilize the same gold ranking label as SLR and optimize the model with RankNet loss, which is the same as what ~\citet{rankgpt} did. We also report the ranking performance of our \textbf{SLR} and \textbf{LLR}, which rerank all the passages using the sliding window strategy.

As for collaborative rerankers, we introduce three naive collaborative baselines, including \textbf{monoT5 (3B) + LLR}, \textbf{RankT5 (3B) + LLR}, and \textbf{Qwen-pointwise (3B) + LLR}. The three baselines first use a corresponding small reranker (3B) to rerank all the passages and then apply LLR to rerank the top-20 passages.

\begin{table*}[]
\centering
\small
\setlength{\tabcolsep}{1.6mm}{
\begin{tabular}{lcccccccccc}
\toprule
 & \multicolumn{1}{c}{} & \multicolumn{2}{c}{TREC DL} & \multicolumn{3}{c}{BEIR} & \multicolumn{3}{c}{BRIGHT} &  \\
 \cmidrule(lr){3-4} \cmidrule(lr){5-7} \cmidrule(lr){8-10}  
Models & Strategy & DL19 & DL20 & Covid & Robust04 & News & Econ. & Earth. & Rob. & Avg. \\ \midrule
BM25 & - & 50.58 & 47.96 & 59.47 & 40.70 & 39.52 & 16.45 & 27.91 & 10.91 & 36.69 \\ \midrule
\multicolumn{11}{c}{Single Reranker} \\ \midrule
monoBERT (340M) & Pointwise & 70.46 & 67.55 & 70.85 & 46.11 & 47.61 & 10.45 & 19.98 & 11.34 & 43.04 \\
monoT5 (3B) & Pointwise & 71.57 & 69.83 & 72.39 & 49.81 & 49.34 & 9.65 & 19.16 & 8.56 & 43.79 \\
RankT5 (3B) & Pointwise & 71.83 & 70.27 & 80.19 & 51.45 & 49.11 & 8.66 & 24.75 & 13.95 & 46.28 \\
Qwen-pointwise (3B) & Pointwise & 70.62 & 64.99 & 81.77 & 53.60 & 47.78 & 14.13 & 26.25 & 9.89 & 46.13 \\
RankVicuna (7B) & Sliding & 67.72 & 65.98 & 79.19 & 48.33 & 47.15 & 11.05 & 16.30 & 8.03 & 42.97 \\
RankZepyer (7B) & Sliding & \textbf{73.39} & 70.02 & {\ul 82.92} & 53.73 & \textbf{52.80} & 15.63 & 18.52 & 12.12 & 47.39 \\
$\text{RankMistral}_{\text{100}}$ (7B) & Full & \multicolumn{1}{l}{72.55} & \multicolumn{1}{l}{\textbf{71.29}} & 82.24 & \textbf{57.91} & 50.59 & {\ul 18.82} & 23.91 & {\ul 15.75} & {\ul 49.13} \\
SLR (3B) & Sliding & 70.24 & 68.85 & 81.85 & 53.68 & 49.85 & 15.56 & \textbf{29.47} & 13.68 & 47.90 \\
LLR (72B) & Sliding & {\ul 73.26} & {\ul 70.51} & \textbf{85.46} & {\ul 57.76} & {\ul 51.28} & \textbf{21.34} & {\ul 28.88} & \textbf{19.04} & \textbf{50.94} \\ \midrule
\multicolumn{11}{c}{Collaborative Rerankers} \\ \midrule
monoT5 (3B) + LLR & Collaborative & 70.79 & {\ul 70.73} & 81.80 & 55.08 & 51.60 & 13.46 & 21.83 & 11.68 & 47.12 \\
RankT5 (3B) + LLR & Collaborative & 69.16 & \textbf{71.41} & {\ul 84.66} & 55.80 & 50.04 & 12.92 & 25.91 & 16.01 & 48.24 \\
Qwen-pointwise (3B) + LLR & Collaborative & {\ul 72.28} & 67.89 & 81.01 & {\ul 56.37} & \textbf{52.30} & {\ul 20.00} & {\ul 28.12} & {\ul 18.26} & {\ul 49.53} \\
\rowcolor{myblue} 
CoRanking & Collaborative & \textbf{72.79} & 70.48 & \textbf{84.90} & \textbf{57.71} & {\ul 52.26} & \textbf{22.72} & \textbf{31.20} & \textbf{19.03} & \textbf{51.39} \\ \bottomrule
\end{tabular}}
\caption{Results (NDCG@10) on TREC, BEIR, and BRIGHT benchmarks. The ``Pointwise'', ``Sliding'', 'Full' and ``Collaborative'' in ``Strategy'' column represent pointwise ranking, sliding window strategy, full ranking~\cite{fullrank} and collaborative ranking, respectively. The best and second-best results for the ``Single Reranker'' part and the ``Collaborative Rerankers'' part are indicated in bold and underlined, respectively. ``Avg.'' represents the averaged result of all 8 datasets.}
\label{tab:main_exp}
\end{table*}

\subsection{Overall Results} \label{subsec:main_results}
The main experimental results are presented in Table~\ref{tab:main_exp}, from which we could draw the following observations:

(1) Our collaborative reranking framework CoRanking achieves the best average performance (51.39) across all datasets, outperforming both single rerankers and other collaborative baselines. Notably, it even surpasses the large listwise reranker LLR by about 0.5 points on average. These results demonstrate that our framework could effectively align the passage order generated by the SLR (3B) with the preferences of the large language model (LLR), thereby achieving better ranking performance than only using LLR.
(2) Regarding single rerankers, our SLR (3B) achieves an average score of 47.90, outperforming all the pointwise rerankers, including monoBERT, monoT5, RankT5, and Qwen-pointwise, and even surpassing 7B-scale RankVicuna and RankZephyr. Furthermore, SLR surpasses Qwen-pointwise by about 1.8 average points under identical training conditions, proving listwise reranking’s superiority through comparison of multiple passages for precise relevance assessment.

(3) Collaborative reranker baselines (\eg, RankT5 (3B) + LLR) improve over corresponding single rerankers (\eg, RankT5 (3B)). However, these baselines still lag behind LLR and our CoRanking, indicating that directly feeding small rerankers’ ranking results to LLR fails to align with LLR's order preference.

\begin{figure}[!t]
        \vspace{-1mm}
        \centering
	\includegraphics[width=0.9\linewidth]{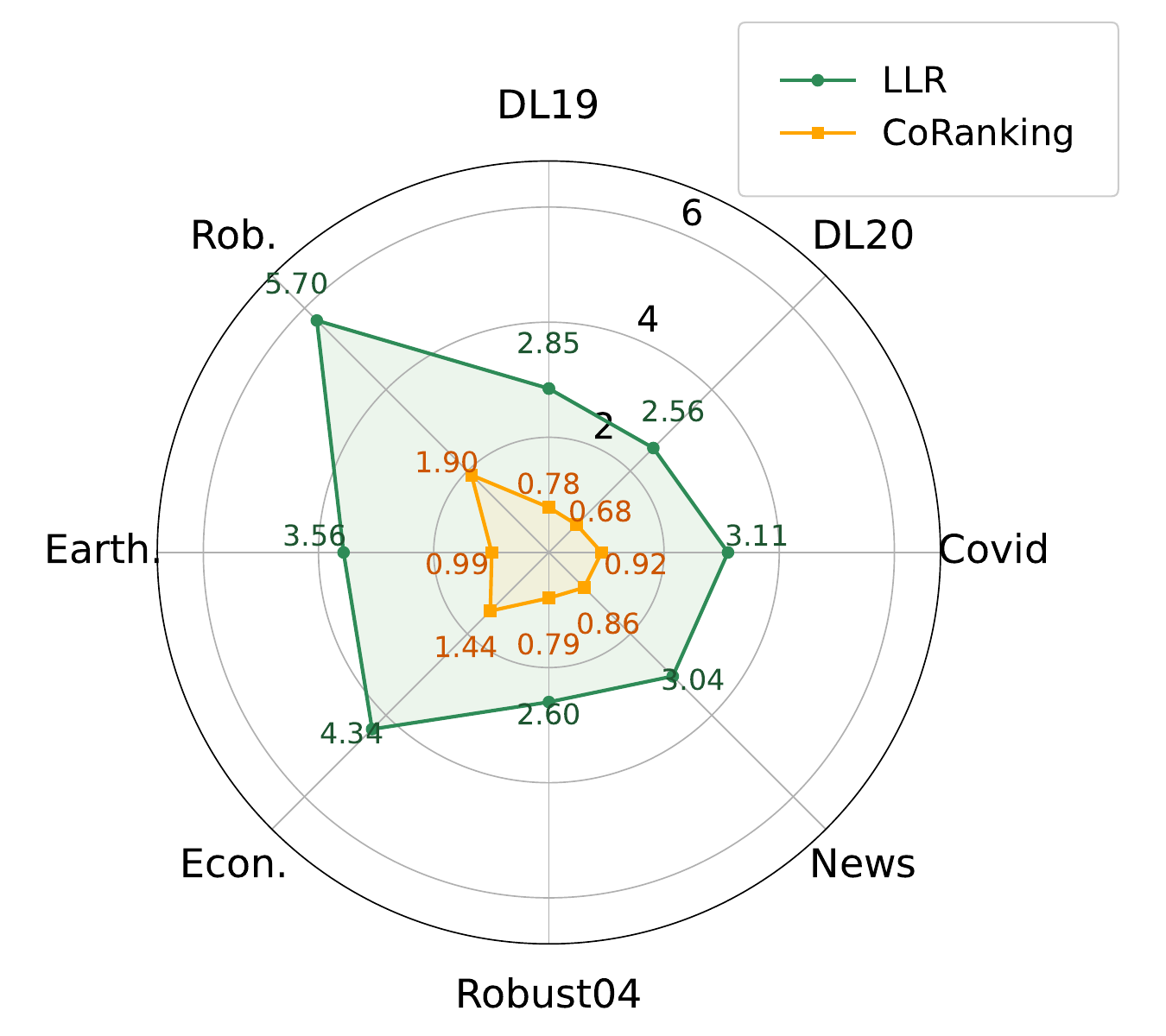}
	\caption{Ranking latency (seconds/query) of LLR and CoRanking on 8 datasets.}
	\label{fig:latency}
\end{figure}

\subsection{Efficiency Analysis} \label{subsec:efficiency}
As mentioned in Section~\ref{sec:intro}, our primary target of CoRanking is to mitigate the efficiency issue of pure LLM-based reranking. To quantitatively demonstrate this advantage, we compare our CoRanking against LLR across all 8 datasets. Our experiments are conducted on 8×40GB NVIDIA A100 GPUs using the vLLM inference framework\footnote{\url{https://github.com/vllm-project/vllm}}. We average across all queries within each dataset and show the comparison results in Figure~\ref{fig:latency}.

Across all datasets, our CoRanking approach demonstrates a significant reduction in latency, achieving a 67\%-74\% decrease compared to the LLR. This substantial improvement highlights the efficiency advantage of CoRanking. Additionally, the latency of less than 1 second achieved by CoRanking across most datasets also showcases its practical efficiency in real-world search engines. Lastly, the ranking latency varies significantly across different datasets, which is primarily due to the differences in passage lengths. This is because longer passages inherently require more computational time for processing.

\begin{table}[]
\centering
\small
\vspace{-2mm}
\setlength{\tabcolsep}{0.7mm}{
\begin{tabular}{lccc}
\toprule
\textbf{Model} & \textbf{TREC Avg.} & \textbf{BEIR Avg.} & \textbf{BRIGHT Avg.} \\ \midrule
CoRanking & \textbf{71.64} & \textbf{64.96} & \textbf{24.32} \\
\hspace{4mm}w/o $\text{S}^3$ strategy & 70.10 & 63.86 & 23.26 \\
\hspace{4mm}w/o POA & 70.23 & 63.80 & 22.74 \\ \midrule
SLR & \textbf{69.55} & \textbf{61.79} & \textbf{19.57} \\
\hspace{4mm}w/o HRC & 68.22 & 59.58 & 19.43 \\ \bottomrule
\end{tabular}}
\caption{Average performance (NDCG@10) of ablated models on TREC (DL19, DL20), BEIR (Covid, Robust04, News), and BRIGHT (Economics, Earth Science, Robotics).}
\label{tab:ablation}
\end{table}

\subsection{Ablation Study} \label{subsec:ablation}
To evaluate the contribution of each component in our framework, we conduct comprehensive ablation experiments. Table~\ref{tab:ablation} presents the average performance for each individual benchmark, with detailed per-dataset results available in Table~\ref{app_tab:ablation}.

First, to evaluate our proposed $\text{S}^3$ sampling strategy for generating the preference pairs of DPO, we introduce a variant, denoted as ``w/o $\text{S}^3$''. This variant uses random sampling for preference pair selection, without ensuring the significant reward of winner prediction and the reward gap between the winner and loser predictions. The performance degradation across all benchmarks (\eg, 1.54-point drop on TREC Avg.) demonstrates the effectiveness of our $\text{S}^3$ sampling in identifying valuable preference pairs for effective DPO training.

We further remove our POA to validate its alignment effectiveness, denoted as ``w/o POA''. In this case, the LLR directly reranks the top-20 passages after SLR's reranking (\ie, the naive collaboration mentioned in Figure~\ref{fig:intro_example} (B)). The notable performance decline (especially 1.6-point decrease on BRIGHT Avg.) demonstrates that POA could effectively align with the order preference of LLR.

Finally, to validate our human-label-enhanced ranking construction (HRC) for fine-tuning SLR, we create a variant (w/o HRC) which generates ranking labels by utilizing LLR to rerank training passages only once and choosing not to use human labels for filtering high-quality ranked lists. The significant performance drops across benchmarks (especially BEIR) highlight HRC's effectiveness in generating high-quality ranking labels.

\begin{figure}[!t]
    \vspace{-3mm}
	\centering
	\includegraphics[width=1\linewidth]{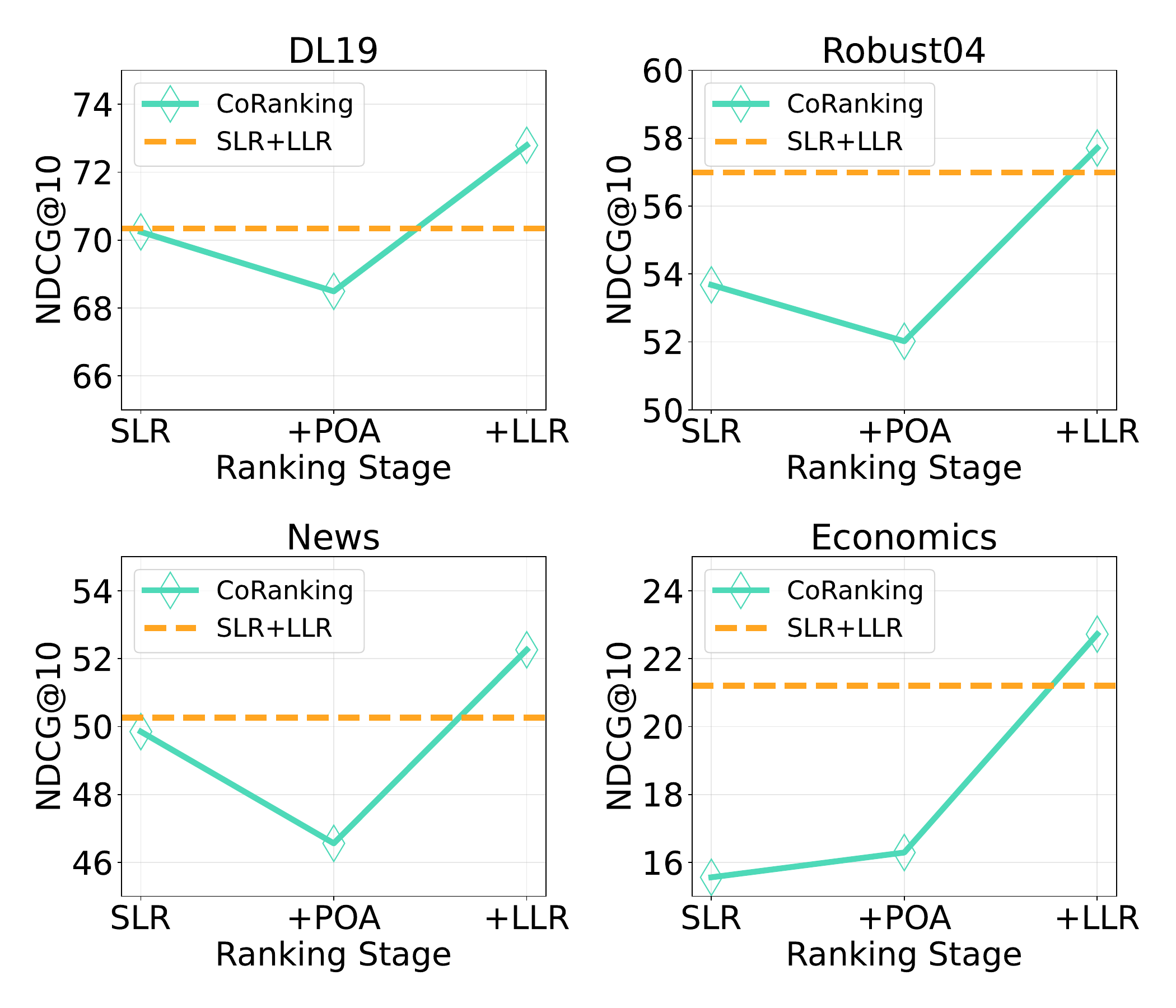}
	\caption{The changes of NDCG@10 after reranking by SLR, POA, and LLR.}
	\label{fig:poa_analysis}
    \vspace{-1mm}
\end{figure}

\subsection{Detailed Analysis of POA} \label{subsec:POA_analysis}
Our POA is designed to adjust the order of the top-20 passages reranked by SLR to align them more with the preferences of the LLR. However, whether this adjustment improves or degrades the ranking quality of top-20 passages remains unknown. We believe this is an interesting question because it helps us understand the reasons why POA works. 

In this part, we selected four datasets—DL19, Robust04, News, and Economics—and reported the changes in the ranking metric NDCG@10 after ranking by SLR, POA, and LLR, respectively. We also report the performance of the naive collaboration SLR+LLR as a baseline. The results are shown in Figure~\ref{fig:poa_analysis}. After POA's adjustment, the ranking metric on three datasets—DL19, Robust04, and News—decreases (except for the Economics dataset). This indicates that a passage ranking list with a high-ranking metric does not necessarily align with LLR’s preferences. This may be due to the positional bias of input tokens during the pre-training of LLMs~\cite{self_consistency_llm}. Moreover, due to the stronger ranking capability, LLR further improves effectiveness compared to SLR and SLR+POA, demonstrating its necessity in the CoRanking framework.

\begin{table}[]
\centering
\small
\setlength{\tabcolsep}{1mm}{
\begin{tabular}{lcccc}
\toprule
\textbf{Model} & \textbf{DL19} & \textbf{News} & \textbf{Earth.} & \textbf{Avg.} \\ \midrule
BM25 & 50.58 & 39.52 & 27.91 & 39.34 \\ \midrule
\multicolumn{5}{c}{Using GPT-4o as LLR} \\ \midrule
RankT5 (3B) + LLR & 73.17 & 49.31 & 29.64 & 50.71 \\
Qwen-pointwise (3B) + LLR & 72.35 & 48.36 & 34.62 & 51.78 \\
SLR+LLR & 73.10 & 48.75 & 34.97 & 52.27 \\
CoRanking & \textbf{74.10} & \textbf{50.19} & \textbf{35.93} & \textbf{53.41} \\ \midrule
\multicolumn{5}{c}{Using DeepSeek-V3 as LLR} \\ \midrule
RankT5 (3B) + LLR & 71.90 & 49.47 & 31.47 & 50.95 \\
Qwen-pointwise (3B) + LLR & 72.16 & 49.07 & 35.38 & 52.20 \\
SLR+LLR & 72.62 & 50.91 & 38.09 & 53.87 \\
CoRanking & \textbf{73.79} & \textbf{51.48} & \textbf{39.01} & \textbf{54.76} \\ \bottomrule
\end{tabular}}
\caption{The performance (NDCG@10) of CoRanking based on different LLRs. ``Avg.'' represents the average performance across the three datasets.}
\label{tab:generalization}
\end{table}

\subsection{Generalization on Other LLMs} \label{subsec:generalization}
In the previous experiments, we use Qwen2.5-72B-Instruct as our LLR model. However, it remains unclear whether our proposed CoRanking framework can generalize effectively to other LLR models. To evaluate this, we conduct additional experiments with two different LLRs: GPT-4o\footnote{The version of GPT-4o we used is gpt-4o-2024-08-06.} and DeepSeek-V3, and compare our CoRanking with three collaborative rerankers (``RankT5 (3B) + LLR'', ``Qwen-pointwise (3B) + LLR'', and ``SLR + LLR''). Note that ``SLR + LLR'' represents the naive collaboration in Figure~\ref{fig:intro_example} (B). For evaluation, we select one dataset from each of the three benchmarks: DL19 from TREC, News from BEIR, and Earth Science from BRIGHT. As illustrated in Table~\ref{tab:generalization}, our CoRanking consistently outperforms the best baseline ``SLR + LLR'' by about 1 point on average, regardless of the LLR used. This demonstrates that the order preference alignment of our POA could generalize to different LLMs.

\section{Related Work}
\subsection{LLMs for Passage Ranking}
The application of large language models (LLMs) into retrieval systems~\cite{llm4ir_survey} has driven substantial research efforts in passage ranking tasks. Current LLM-based ranking methods can be categorized into three categories: pointwise, pairwise, and listwise. Pointwise methods~\cite{liang2022holistic, sachan2022improving, beyond_yes_no, demorank, LiuZZD24} compute the relevance score for each query-passage pair. Pairwise methods~\cite{qin2023large, PRP-Graph} compare two passages at a time and identify the more relevant one. Listwise methods~\cite{rankgpt, rankvicuna, first, fullrank, pe_rank, tourrank} directly take a list of passages as the input and output the passage identifiers in descending order of their relevance. Through comparing multiple passages, a listwise reranker could assess the relevance more comprehensively and yield better ranking results. Due to the superior performance of listwise ranking, it has attracted the interest of many researchers~\cite{self_consistency_llm, rankzephyr, liu2025reasonrank, fan2025llm}.


The superior performance of LLM listwise rerankers is often attributed to the utilization of large-scale LLMs and the sliding window strategy, which leads to an efficiency issue in ranking. In this paper, we introduce a novel framework that leverages the collaborative ranking of large rerankers and small rerankers to enhance the ranking efficiency while maintaining ranking effectiveness.

\subsection{Preference Alignment}
Preference alignment has emerged as a critical research direction, aiming to align model behaviors with human or task-specific preferences. Early approaches primarily relied on reinforcement learning from human feedback (RLHF), where models are fine-tuned using reward signals derived from human preference annotations. To reduce the computational complexity of RLHF, recent studies propose more efficient alternatives, such as DPO~\cite{dpo} and Step-DPO~\cite{stepdpo}. Direct preference optimization (DPO)~\cite{dpo} reformulates preference learning as a supervised objective, bypassing explicit reward modeling while achieving competitive performance. 
In this paper, to address the positional sensitivity of a large listwise reranker, we propose training a passage order adjuster with the DPO algorithm to align the passage order generated by the small listwise reranker with the order preference of the large listwise reranker.

\section{Conclusion}
In this paper, we introduce a novel collaborative ranking framework, namely CoRanking, to combine small and large listwise rerankers for more efficient and effective passage ranking. Specifically, we first train a small listwise reranker SLR to move the relevant passages to the top of the list. Then we introduce a passage order adjuster POA to rerank the top-ranked passages to better align with the order preference of a large listwise reranker. Finally, the large listwise reranker is utilized to solely rerank the top passages. Extensive experiments demonstrate the efficiency advantage and superior performance of our CoRanking.

\section*{Limitations} \label{limitation}
We acknowledge some limitations in our work. First, our framework consists of three models, making its online deployment more complex compared to using a single unified reranker. Second, we only use Qwen2.5-3B-Instruct as the backbone model of our SLR and POA. We believe it is meaningful to attempt using different backbones (such as Llama-series models) and models of varying sizes (such as Qwen2.5-1.5B-Instruct) to further demonstrate the generalization of our CoRanking framework.

\section*{Acknowledgments}
This work was supported by National Natural Science Foundation of China No. 62272467, Beijing Municipal Science and Technology Project No. Z231100010323009, Beijing Natural Science Foundation No. L233008. The work was partially done at the Engineering Research Center of Next-Generation Intelligent Search and Recommendation, MOE.


\begin{thebibliography}{32}
\expandafter\ifx\csname natexlab\endcsname\relax\def\natexlab#1{#1}\fi

\bibitem[{Chen et~al.(2024)Chen, Liu, Zhang, Sun, Shi, Mao, and Yin}]{tourrank}
Yiqun Chen, Qi~Liu, Yi~Zhang, Weiwei Sun, Daiting Shi, Jiaxin Mao, and Dawei Yin. 2024.
\newblock Tourrank: Utilizing large language models for documents ranking with a tournament-inspired strategy.
\newblock \emph{CoRR}, abs/2406.11678.

\bibitem[{Craswell et~al.(2020{\natexlab{a}})Craswell, Mitra, Yilmaz, and Campos}]{dl20}
Nick Craswell, Bhaskar Mitra, Emine Yilmaz, and Daniel Campos. 2020{\natexlab{a}}.
\newblock Overview of the {TREC} 2020 deep learning track.
\newblock In \emph{{TREC}}, volume 1266 of \emph{{NIST} Special Publication}. National Institute of Standards and Technology {(NIST)}.

\bibitem[{Craswell et~al.(2020{\natexlab{b}})Craswell, Mitra, Yilmaz, Campos, and Voorhees}]{dl19}
Nick Craswell, Bhaskar Mitra, Emine Yilmaz, Daniel Campos, and Ellen~M. Voorhees. 2020{\natexlab{b}}.
\newblock Overview of the {TREC} 2019 deep learning track.
\newblock \emph{CoRR}, abs/2003.07820.

\bibitem[{Fan et~al.(2025)Fan, Xue, Li, Zhang, and Ruan}]{fan2025llm}
Yongqi Fan, Kui Xue, Zelin Li, Xiaofan Zhang, and Tong Ruan. 2025.
\newblock An llm-based framework for biomedical terminology normalization in social media via multi-agent collaboration.
\newblock In \emph{Proceedings of the 31st International Conference on Computational Linguistics}, pages 10712--10726.

\bibitem[{Lai et~al.(2024)Lai, Tian, Chen, Yang, Peng, and Jia}]{stepdpo}
Xin Lai, Zhuotao Tian, Yukang Chen, Senqiao Yang, Xiangru Peng, and Jiaya Jia. 2024.
\newblock \href {https://doi.org/10.48550/ARXIV.2406.18629} {Step-dpo: Step-wise preference optimization for long-chain reasoning of llms}.
\newblock \emph{CoRR}, abs/2406.18629.

\bibitem[{Liang et~al.(2022)Liang, Bommasani, Lee, Tsipras, Soylu, Yasunaga, Zhang, Narayanan, Wu, Kumar, Newman, Yuan, Yan, Zhang, Cosgrove, Manning, R{\'{e}}, Acosta{-}Navas, Hudson, Zelikman, Durmus, Ladhak, Rong, Ren, Yao, Wang, Santhanam, Orr, Zheng, Y{\"{u}}ksekg{\"{o}}n{\"{u}}l, Suzgun, Kim, Guha, Chatterji, Khattab, Henderson, Huang, Chi, Xie, Santurkar, Ganguli, Hashimoto, Icard, Zhang, Chaudhary, Wang, Li, Mai, Zhang, and Koreeda}]{liang2022holistic}
Percy Liang, Rishi Bommasani, Tony Lee, Dimitris Tsipras, Dilara Soylu, Michihiro Yasunaga, Yian Zhang, Deepak Narayanan, Yuhuai Wu, Ananya Kumar, Benjamin Newman, Binhang Yuan, Bobby Yan, Ce~Zhang, Christian Cosgrove, Christopher~D. Manning, Christopher R{\'{e}}, Diana Acosta{-}Navas, Drew~A. Hudson, Eric Zelikman, Esin Durmus, Faisal Ladhak, Frieda Rong, Hongyu Ren, Huaxiu Yao, Jue Wang, Keshav Santhanam, Laurel~J. Orr, Lucia Zheng, Mert Y{\"{u}}ksekg{\"{o}}n{\"{u}}l, Mirac Suzgun, Nathan Kim, Neel Guha, Niladri~S. Chatterji, Omar Khattab, Peter Henderson, Qian Huang, Ryan Chi, Sang~Michael Xie, Shibani Santurkar, Surya Ganguli, Tatsunori Hashimoto, Thomas Icard, Tianyi Zhang, Vishrav Chaudhary, William Wang, Xuechen Li, Yifan Mai, Yuhui Zhang, and Yuta Koreeda. 2022.
\newblock \href {https://doi.org/10.48550/arXiv.2211.09110} {Holistic evaluation of language models}.
\newblock \emph{CoRR}, abs/2211.09110.

\bibitem[{Liu et~al.(2024{\natexlab{a}})Liu, Lin, Hewitt, Paranjape, Bevilacqua, Petroni, and Liang}]{LiuLHPBPL24}
Nelson~F. Liu, Kevin Lin, John Hewitt, Ashwin Paranjape, Michele Bevilacqua, Fabio Petroni, and Percy Liang. 2024{\natexlab{a}}.
\newblock \href {https://doi.org/10.1162/TACL\_A\_00638} {Lost in the middle: How language models use long contexts}.
\newblock \emph{Trans. Assoc. Comput. Linguistics}, 12:157--173.

\bibitem[{Liu et~al.(2024{\natexlab{b}})Liu, Wang, Wang, and Mao}]{pe_rank}
Qi~Liu, Bo~Wang, Nan Wang, and Jiaxin Mao. 2024{\natexlab{b}}.
\newblock Leveraging passage embeddings for efficient listwise reranking with large language models.
\newblock \emph{CoRR}, abs/2406.14848.

\bibitem[{Liu et~al.(2025)Liu, Ma, Sun, Zhu, Li, Yin, and Dou}]{liu2025reasonrank}
Wenhan Liu, Xinyu Ma, Weiwei Sun, Yutao Zhu, Yuchen Li, Dawei Yin, and Zhicheng Dou. 2025.
\newblock Reasonrank: Empowering passage ranking with strong reasoning ability.
\newblock \emph{arXiv preprint arXiv:2508.07050}.

\bibitem[{Liu et~al.(2024{\natexlab{c}})Liu, Ma, Zhu, Zhao, Wang, Yin, and Dou}]{fullrank}
Wenhan Liu, Xinyu Ma, Yutao Zhu, Ziliang Zhao, Shuaiqiang Wang, Dawei Yin, and Zhicheng Dou. 2024{\natexlab{c}}.
\newblock \href {https://doi.org/10.48550/ARXIV.2412.14574} {Sliding windows are not the end: Exploring full ranking with long-context large language models}.
\newblock \emph{CoRR}, abs/2412.14574.

\bibitem[{Liu et~al.(2024{\natexlab{d}})Liu, Zhou, Zhu, and Dou}]{LiuZZD24}
Wenhan Liu, Yujia Zhou, Yutao Zhu, and Zhicheng Dou. 2024{\natexlab{d}}.
\newblock \href {https://doi.org/10.1007/S10115-024-02138-Y} {How to personalize and whether to personalize? candidate documents decide}.
\newblock \emph{Knowl. Inf. Syst.}, 66(9):5581--5604.

\bibitem[{Liu et~al.(2024{\natexlab{e}})Liu, Zhu, and Dou}]{demorank}
Wenhan Liu, Yutao Zhu, and Zhicheng Dou. 2024{\natexlab{e}}.
\newblock Demorank: Selecting effective demonstrations for large language models in ranking task.
\newblock \emph{CoRR}, abs/2406.16332.

\bibitem[{Luo et~al.(2024)Luo, Chen, He, and Sun}]{PRP-Graph}
Jian Luo, Xuanang Chen, Ben He, and Le~Sun. 2024.
\newblock Prp-graph: Pairwise ranking prompting to llms with graph aggregation for effective text re-ranking.
\newblock In \emph{{ACL} {(1)}}, pages 5766--5776. Association for Computational Linguistics.

\bibitem[{Nguyen et~al.(2016)Nguyen, Rosenberg, Song, Gao, Tiwary, Majumder, and Deng}]{MSMARCO}
Tri Nguyen, Mir Rosenberg, Xia Song, Jianfeng Gao, Saurabh Tiwary, Rangan Majumder, and Li~Deng. 2016.
\newblock \href {https://ceur-ws.org/Vol-1773/CoCoNIPS\_2016\_paper9.pdf} {{MS} {MARCO:} {A} human generated machine reading comprehension dataset}.
\newblock In \emph{Proceedings of the Workshop on Cognitive Computation: Integrating neural and symbolic approaches 2016 co-located with the 30th Annual Conference on Neural Information Processing Systems {(NIPS} 2016), Barcelona, Spain, December 9, 2016}, volume 1773 of \emph{{CEUR} Workshop Proceedings}. CEUR-WS.org.

\bibitem[{Nogueira and Cho(2019)}]{monobert}
Rodrigo~Frassetto Nogueira and Kyunghyun Cho. 2019.
\newblock \href {http://arxiv.org/abs/1901.04085} {Passage re-ranking with {BERT}}.
\newblock \emph{CoRR}, abs/1901.04085.

\bibitem[{Nogueira et~al.(2020)Nogueira, Jiang, Pradeep, and Lin}]{monot5}
Rodrigo~Frassetto Nogueira, Zhiying Jiang, Ronak Pradeep, and Jimmy Lin. 2020.
\newblock \href {https://doi.org/10.18653/v1/2020.findings-emnlp.63} {Document ranking with a pretrained sequence-to-sequence model}.
\newblock In \emph{Findings of the Association for Computational Linguistics: {EMNLP} 2020, Online Event, 16-20 November 2020}, volume {EMNLP} 2020 of \emph{Findings of {ACL}}, pages 708--718. Association for Computational Linguistics.

\bibitem[{Pradeep et~al.(2023{\natexlab{a}})Pradeep, Sharifymoghaddam, and Lin}]{rankvicuna}
Ronak Pradeep, Sahel Sharifymoghaddam, and Jimmy Lin. 2023{\natexlab{a}}.
\newblock \href {https://doi.org/10.48550/arXiv.2309.15088} {Rankvicuna: Zero-shot listwise document reranking with open-source large language models}.
\newblock \emph{CoRR}, abs/2309.15088.

\bibitem[{Pradeep et~al.(2023{\natexlab{b}})Pradeep, Sharifymoghaddam, and Lin}]{rankzephyr}
Ronak Pradeep, Sahel Sharifymoghaddam, and Jimmy Lin. 2023{\natexlab{b}}.
\newblock Rankzephyr: Effective and robust zero-shot listwise reranking is a breeze!
\newblock \emph{CoRR}, abs/2312.02724.

\bibitem[{Qin et~al.(2023)Qin, Jagerman, Hui, Zhuang, Wu, Shen, Liu, Liu, Metzler, Wang, and Bendersky}]{qin2023large}
Zhen Qin, Rolf Jagerman, Kai Hui, Honglei Zhuang, Junru Wu, Jiaming Shen, Tianqi Liu, Jialu Liu, Donald Metzler, Xuanhui Wang, and Michael Bendersky. 2023.
\newblock \href {https://doi.org/10.48550/arXiv.2306.17563} {Large language models are effective text rankers with pairwise ranking prompting}.
\newblock \emph{CoRR}, abs/2306.17563.

\bibitem[{Rafailov et~al.(2023)Rafailov, Sharma, Mitchell, Manning, Ermon, and Finn}]{dpo}
Rafael Rafailov, Archit Sharma, Eric Mitchell, Christopher~D. Manning, Stefano Ermon, and Chelsea Finn. 2023.
\newblock \href {http://papers.nips.cc/paper\_files/paper/2023/hash/a85b405ed65c6477a4fe8302b5e06ce7-Abstract-Conference.html} {Direct preference optimization: Your language model is secretly a reward model}.
\newblock In \emph{Advances in Neural Information Processing Systems 36: Annual Conference on Neural Information Processing Systems 2023, NeurIPS 2023, New Orleans, LA, USA, December 10 - 16, 2023}.

\bibitem[{Reddy et~al.(2024)Reddy, Doo, Xu, Sultan, Swain, Sil, and Ji}]{first}
Revanth~Gangi Reddy, JaeHyeok Doo, Yifei Xu, Md.~Arafat Sultan, Deevya Swain, Avirup Sil, and Heng Ji. 2024.
\newblock {FIRST:} faster improved listwise reranking with single token decoding.
\newblock \emph{CoRR}, abs/2406.15657.

\bibitem[{Sachan et~al.(2022)Sachan, Lewis, Joshi, Aghajanyan, Yih, Pineau, and Zettlemoyer}]{sachan2022improving}
Devendra~Singh Sachan, Mike Lewis, Mandar Joshi, Armen Aghajanyan, Wen{-}tau Yih, Joelle Pineau, and Luke Zettlemoyer. 2022.
\newblock \href {https://doi.org/10.18653/v1/2022.emnlp-main.249} {Improving passage retrieval with zero-shot question generation}.
\newblock In \emph{Proceedings of the 2022 Conference on Empirical Methods in Natural Language Processing, {EMNLP} 2022, Abu Dhabi, United Arab Emirates, December 7-11, 2022}, pages 3781--3797. Association for Computational Linguistics.

\bibitem[{Soboroff et~al.(2019)Soboroff, Huang, and Harman}]{news}
Ian Soboroff, Shudong Huang, and Donna Harman. 2019.
\newblock Trec 2019 news track overview.
\newblock In \emph{TREC}.

\bibitem[{Su et~al.(2024)Su, Yen, Xia, Shi, Muennighoff, Wang, Liu, Shi, Siegel, Tang, Sun, Yoon, Arik, Chen, and Yu}]{bright}
Hongjin Su, Howard Yen, Mengzhou Xia, Weijia Shi, Niklas Muennighoff, Han{-}yu Wang, Haisu Liu, Quan Shi, Zachary~S. Siegel, Michael Tang, Ruoxi Sun, Jinsung Yoon, Sercan~{\"{O}}. Arik, Danqi Chen, and Tao Yu. 2024.
\newblock \href {https://doi.org/10.48550/ARXIV.2407.12883} {{BRIGHT:} {A} realistic and challenging benchmark for reasoning-intensive retrieval}.
\newblock \emph{CoRR}, abs/2407.12883.

\bibitem[{Sun et~al.(2023)Sun, Yan, Ma, Wang, Ren, Chen, Yin, and Ren}]{rankgpt}
Weiwei Sun, Lingyong Yan, Xinyu Ma, Shuaiqiang Wang, Pengjie Ren, Zhumin Chen, Dawei Yin, and Zhaochun Ren. 2023.
\newblock \href {https://aclanthology.org/2023.emnlp-main.923} {Is chatgpt good at search? investigating large language models as re-ranking agents}.
\newblock In \emph{Proceedings of the 2023 Conference on Empirical Methods in Natural Language Processing, {EMNLP} 2023, Singapore, December 6-10, 2023}, pages 14918--14937. Association for Computational Linguistics.

\bibitem[{Tang et~al.(2024)Tang, Zhang, Ma, Lin, and Ture}]{self_consistency_llm}
Raphael Tang, Xinyu~Crystina Zhang, Xueguang Ma, Jimmy Lin, and Ferhan Ture. 2024.
\newblock \href {https://doi.org/10.18653/V1/2024.NAACL-LONG.129} {Found in the middle: Permutation self-consistency improves listwise ranking in large language models}.
\newblock In \emph{Proceedings of the 2024 Conference of the North American Chapter of the Association for Computational Linguistics: Human Language Technologies (Volume 1: Long Papers), {NAACL} 2024, Mexico City, Mexico, June 16-21, 2024}, pages 2327--2340. Association for Computational Linguistics.

\bibitem[{Thakur et~al.(2021)Thakur, Reimers, R{\"{u}}ckl{\'{e}}, Srivastava, and Gurevych}]{beir}
Nandan Thakur, Nils Reimers, Andreas R{\"{u}}ckl{\'{e}}, Abhishek Srivastava, and Iryna Gurevych. 2021.
\newblock {BEIR:} {A} heterogeneous benchmark for zero-shot evaluation of information retrieval models.
\newblock In \emph{NeurIPS Datasets and Benchmarks}.

\bibitem[{Voorhees(2004)}]{robust04}
Ellen~M. Voorhees. 2004.
\newblock \href {http://trec.nist.gov/pubs/trec13/papers/ROBUST.OVERVIEW.pdf} {Overview of the {TREC} 2004 robust track}.
\newblock In \emph{Proceedings of the Thirteenth Text REtrieval Conference, {TREC} 2004, Gaithersburg, Maryland, USA, November 16-19, 2004}, volume 500-261 of \emph{{NIST} Special Publication}. National Institute of Standards and Technology {(NIST)}.

\bibitem[{Voorhees et~al.(2020)Voorhees, Alam, Bedrick, Demner{-}Fushman, Hersh, Lo, Roberts, Soboroff, and Wang}]{covid}
Ellen~M. Voorhees, Tasmeer Alam, Steven Bedrick, Dina Demner{-}Fushman, William~R. Hersh, Kyle Lo, Kirk Roberts, Ian Soboroff, and Lucy~Lu Wang. 2020.
\newblock \href {https://doi.org/10.1145/3451964.3451965} {{TREC-COVID:} constructing a pandemic information retrieval test collection}.
\newblock \emph{{SIGIR} Forum}, 54(1):1:1--1:12.

\bibitem[{Zhu et~al.(2023)Zhu, Yuan, Wang, Liu, Liu, Deng, Dou, and Wen}]{llm4ir_survey}
Yutao Zhu, Huaying Yuan, Shuting Wang, Jiongnan Liu, Wenhan Liu, Chenlong Deng, Zhicheng Dou, and Ji{-}Rong Wen. 2023.
\newblock \href {https://doi.org/10.48550/arXiv.2308.07107} {Large language models for information retrieval: {A} survey}.
\newblock \emph{CoRR}, abs/2308.07107.

\bibitem[{Zhuang et~al.(2023{\natexlab{a}})Zhuang, Qin, Hui, Wu, Yan, Wang, and Bendersky}]{beyond_yes_no}
Honglei Zhuang, Zhen Qin, Kai Hui, Junru Wu, Le~Yan, Xuanhui Wang, and Michael Bendersky. 2023{\natexlab{a}}.
\newblock \href {https://doi.org/10.48550/ARXIV.2310.14122} {Beyond yes and no: Improving zero-shot {LLM} rankers via scoring fine-grained relevance labels}.
\newblock \emph{CoRR}, abs/2310.14122.

\bibitem[{Zhuang et~al.(2023{\natexlab{b}})Zhuang, Qin, Jagerman, Hui, Ma, Lu, Ni, Wang, and Bendersky}]{rankt5}
Honglei Zhuang, Zhen Qin, Rolf Jagerman, Kai Hui, Ji~Ma, Jing Lu, Jianmo Ni, Xuanhui Wang, and Michael Bendersky. 2023{\natexlab{b}}.
\newblock Rankt5: Fine-tuning {T5} for text ranking with ranking losses.
\newblock In \emph{Proceedings of the 46th International {ACM} {SIGIR} Conference on Research and Development in Information Retrieval, {SIGIR} 2023, Taipei, Taiwan, July 23-27, 2023}, pages 2308--2313. {ACM}.

\end{thebibliography}

\clearpage
\appendix

\section{Listwise Ranking Prompt} \label{app:listwise_prompt}
The listwise ranking prompt of LLR used in this paper is shown as bellow:
\begin{tcolorbox}[colback=gray!10, colframe=black, title=Prompt: Listwise Ranking Prompt]
You are RankLLM, an intelligent assistant that can rank passages based on their relevancy to the query.

I will provide you with \texttt{\{num\}} passages, each indicated by a numerical identifier \texttt{[]}.

Rank the passages based on their relevance to the search query: \texttt{\{query\}}.

\texttt{[1]} \{passage 1\}  

\texttt{[2]} \{passage 2\}

...

\texttt{[\{num\}]} \{passage \{num\}\} \\

Search Query: \texttt{\{query\}}. \\

Rank the \texttt{\{num\}} passages above based on their relevance to the search query. All the passages should be included and listed using identifiers, in descending order of relevance. The output format should be \texttt{[]} > \texttt{[]}, e.g., \texttt{[4]} > \texttt{[2]}. Only respond with the ranking results, do not say any word or explain.
\end{tcolorbox}

\section{Datasets Details} \label{app:datasets}
In this part, we provide the details of each evaluation dataset we used:

\noindent$\bullet$ \textbf{DL19/DL20}
The TREC Deep Learning 2019 (DL19)~\cite{dl19} and 2020 (DL20)~\cite{dl20} datasets are derived from the MS MARCO V1 passage ranking corpus. DL19 contains 43 test queries with dense relevance annotations from the TREC 2019 Deep Learning Track, while DL20 includes 54 queries from the 2020 edition. Both datasets focus on web search scenarios and are widely adopted as standard benchmarks for large-scale information retrieval systems.

\noindent$\bullet$ \textbf{TREC-Covid}~\cite{covid}: A biomedical dataset comprising scientific articles related to COVID-19, with 50 test queries from the TREC 2020 COVID Track.

\noindent$\bullet$\textbf{Robust04}~\cite{robust04}: A news retrieval dataset from the TREC Robust Track 2004, featuring 249 queries with ambiguous or complex information needs.

\noindent$\bullet$\textbf{TREC-News}~\cite{news}: A web-archive retrieval dataset containing news articles from the Washington Post corpus, evaluated with 57 test queries requiring event-oriented semantic matching.

\noindent$\bullet$\textbf{Economics}: Focuses on technical questions in economics, including policy analysis and theoretical discussions from StackExchange. Contains 103 complex queries derived from user posts.

\noindent$\bullet$\textbf{Earth Science}: Targets geophysical and climate-related inquiries, such as weather patterns and environmental processes. Includes 116 queries based on detailed observational questions.

\noindent$\bullet$\textbf{Robotics}: Centers on robotics engineering challenges, including system errors and design optimizations. Comprises 101 queries from StackExchange’s technical discussions.


\begin{table*}[]
\centering
\small
\setlength{\tabcolsep}{2.7mm}{
\begin{tabular}{lccccccccc}
\toprule
 & \multicolumn{2}{c}{TREC} & \multicolumn{3}{c}{BEIR} & \multicolumn{3}{c}{BRIGHT} &  \\
  \cmidrule(lr){2-3} \cmidrule(lr){4-6} \cmidrule(lr){7-9}  
Model & DL19 & DL20 & Covid & Robust04 & News & Econ. & Earth. & Rob. & Avg. \\ \midrule
CoRanking & \textbf{72.79} & \textbf{70.48} & \textbf{84.90} & \textbf{57.71} & \textbf{52.26} & \textbf{22.72} & \textbf{31.20} & \textbf{19.03} & \textbf{51.39} \\
\hspace{4mm}w/o $\text{S}^3$ strategy & 70.90 & 69.30 & 83.18 & 56.91 & 51.48 & 21.88 & 30.38 & 17.51 & 50.19 \\
\hspace{4mm}w/o POA & 70.34 & 70.11 & 84.14 & 56.99 & 50.26 & 21.20 & 30.50 & 16.51 & 50.01 \\ \midrule
SLR & \textbf{70.24} & \textbf{68.85} & \textbf{81.85} & \textbf{53.68} & \textbf{49.85} & 15.56 & 29.47 & \textbf{13.68} & \textbf{47.90} \\
\hspace{4mm}w/o HRC & 69.53 & 66.90 & 79.36 & 50.61 & 48.77 & \textbf{15.59} & \textbf{29.66} & 13.04 & 46.68 \\ \bottomrule
\end{tabular}}
\caption{Results (NDCG@10) of ablated models on TREC, BEIR, and BRIGHT benchmarks. The best-performing models are marked in bold.}
\label{app_tab:ablation}
\end{table*}

\section{Baselines Details} \label{app:baselines}
In this paper, we include two kinds of baselines: (1) Single reranker and (2) Collaborative rerankers. The details of single reranker are shown as below:
\noindent$\bullet$ \textbf{monoBERT (340M)}~\cite{monobert}: A BERT-large-based cross-encoder re-ranker.

\noindent$\bullet$ \textbf{monoT5 (3B)}~\cite{monot5}: A T5-based sequence-to-sequence re-ranker that computes relevance scores via pointwise ranking loss.

\noindent$\bullet$ \textbf{RankT5 (3B)}~\cite{rankt5}: A T5-based re-ranker optimized with listwise ranking loss.

\noindent$\bullet$ \textbf{Qwen-pointwise (3B)}: To ensure fair comparison in model architecture and training data, we fine-tune a pointwise re-ranker based on Qwen2.5-3B-Instruct. This model concatenates queries and passages as input to predict relevance scores, trained on SLR data with RankNet loss following~\cite{rankgpt}.

\noindent$\bullet$ \textbf{RankVicuna}~\cite{rankvicuna}: RankVicuna is a listwise re-ranker distilled from GPT-3.5-generated ranked lists.

\noindent$\bullet$ \textbf{RankZephyr}~\cite{rankzephyr}: RankZephyr is a listwise reranker distilled from GPT-3.5 and GPT-4.

\noindent$\bullet$ $\textbf{RankMistral}_{\text{100}}$~\cite{fullrank}: $\textbf{RankMistral}_{\text{100}}$ is a full reranker trained using a full ranking list generated by GPT-4o. It directly takes all the passages as input and output the reranked list without relying on a sliding window strategy.

\noindent$\bullet$ \textbf{LLR (72B)}: The large listwise reranker that employs a zero-shot sliding window strategy for passage re-ranking.

\section{More Implementation Details} \label{app:implementation}
Unlike previous studies~\cite{rankzephyr, fullrank} that adopted GPT-series models (\eg, GPT-4o) as the teacher model, we select Qwen2.5-72B-Instruct as it demonstrates comparable and even better performance to GPT-4o across most datasets while being more cost-effective. Nevertheless, we have included additional experimental results using GPT-4o as the LLR model in Section~\ref{subsec:generalization}. We train the SLR for 4 epochs with a learning rate of 5e-6 and a batch size of 1. The batch size, learning rate, and the hyperparameter $\beta$ of DPO are set as 1, 1e-6, and 0.4, respectively.

\end{document}